\documentclass[runningheads]{llncs}
\usepackage{graphicx}
\usepackage{amsmath,amssymb} 
\usepackage{color}
\usepackage{subfig}
\usepackage{multicol}
\usepackage{multirow}
\usepackage[labelfont=bf]{caption}
\begin{document}

\newcommand{\point}{
    \raise0.7ex\hbox{.}
    }


\pagestyle{headings}

\mainmatter

\title{Deep Supervised Hashing with Triplet Labels} 

\titlerunning{Deep Supervised Hashing with Triplet Labels} 

\authorrunning{Xiaofang Wang, Yi Shi, Kris M. Kitani} 

\author{Xiaofang Wang, Yi Shi, Kris M. Kitani} 
\institute{Carnegie Mellon University, Pittsburgh, PA 15213 USA} 

\maketitle

\begin{abstract}
Hashing is one of the most popular and powerful approximate nearest neighbor search techniques for large-scale image retrieval. Most traditional hashing methods first represent images as off-the-shelf visual features and then produce hashing codes in a separate stage. However, off-the-shelf visual features may not be optimally compatible with the hash code learning procedure, which may result in sub-optimal hash codes. Recently, deep hashing methods have been proposed to simultaneously learn image features and hash codes using deep neural networks and have shown superior performance over traditional hashing methods. Most deep hashing methods are given supervised information in the form of pairwise labels or triplet labels.  The current state-of-the-art deep hashing method DPSH~\cite{li2015feature}, which is based on pairwise labels, performs image feature learning and hash code learning simultaneously by maximizing the likelihood of pairwise similarities. Inspired by DPSH~\cite{li2015feature}, we propose a triplet label based deep hashing method which aims to maximize the likelihood of the given triplet labels. Experimental results show that our method outperforms all the baselines on CIFAR-10 and NUS-WIDE datasets, including the state-of-the-art method DPSH~\cite{li2015feature} and all the previous triplet label based deep hashing methods.
\end{abstract}

\section{Introduction}

With the rapid growth of image data on the Internet, much attention has been devoted to approximate nearest neighbor (ANN) search. Hashing is one of the most popular and powerful techniques for ANN search due to its computational and storage efficiencies. Hashing aims to map high dimensional image features into compact hash codes or binary codes so that the Hamming distance between hash codes approximates the Euclidean distance between image features.

Many hashing methods have been proposed and they can be categorized into data-independent and data-dependent methods. Compared with the data-dependent methods, data-independent methods need longer codes to achieve satisfactory performance~\cite{gong2011iterative}. Data-dependent methods can be further categorized into unsupervised and supervised methods. Compared to unsupervised methods, supervised methods usually can achieve competitive performance with fewer bits due to the help of supervised information, which is advantageous for search speed and storage efficiency~\cite{lai2015simultaneous}.

Most existing hashing methods first represent images as off-the-shelf visual features such as GIST~\cite{oliva2001modeling}, SIFT~\cite{lowe2004distinctive} and hash code learning procedure is independent of the features of images. However, off-the-shelf visual features may not be optimally compatible with hash code learning procedure. In other words, the similarity between two images may not be optimally preserved by the visual features and thus the learned hash codes are sub-optimal~\cite{lai2015simultaneous}. Therefore, those hashing methods may not be able to achieve satisfactory performance in practice. To address the drawbacks of hashing methods that rely on off-the-shelf visual features, feature learning based deep hashing methods~\cite{zhao2015deep,zhang2015bit,lai2015simultaneous,li2015feature} have been proposed to simultaneously learn image feature and hash codes with deep neural networks and have demonstrated superior performance over traditional hashing methods. Most proposed deep hashing methods fall into the category of supervised hashing methods. Supervised information is given in the form of pairwise labels or triplet labels, a special case of ranking labels.

DPSH~\cite{li2015feature} is the current state-of-the-art deep hashing method, which is supervised by pairwise labels. Similar to LFH~\cite{zhang2015bit}, DPSH aims to maximize the likelihood of the pairwise similarities, which is modeled as a function of the Hamming distance between the corresponding data points.
\par
We argue that triplet labels inherently contain richer information than pairwise labels. Each triplet label can be naturally decomposed into two pairwise labels. Whereas, a triplet label can be constructed from two pairwise labels only when the same query image is used in a positive pairwise label and a negative pairwise label simultaneously. A triplet label ensures that in the learned hash code space, the query image is close to the positive image and far from the negative image simultaneously. However, a pairwise label can only ensure that one constraint is observed. Triplet labels explicitly provide a notion of relative similarities between images while pairwise labels can only encode that implicitly.
\par
Therefore, we propose a triplet label based deep hashing method, which performs image  feature  learning  and  hash code learning simultaneously by maximizing the likelihood of the given triplet labels. As shown in Fig.~\ref{framework}, the proposed model has three key components: (1) image feature learning component: a deep neural network to learn visual features from images, (2) hash code learning component: one fully connected layer to learn hash codes from image features and (3) loss function component: a loss function to measure how well the given triplet labels are satisfied by the learned hash codes by computing the likelihood of the given triplet labels. Extensive experiments on standard benchmark datasets such as CIFAR-10 and NUS-WIDE show that our proposed deep hashing method  outperforms all the baselines, including the state-of-the-art method DPSH~\cite{li2015feature} and all the previous triplet label based deep hashing methods.

\par{\textbf{Contributions}}: (1) We propose a novel triplet label based deep hashing method to simultaneously perform image feature and hash code learning in an end-to-end manner; (2) We present a novel formulation of the likelihood of the given triplet labels to evaluate the quality of learned hash codes; (3) We provide ablative analysis of our loss function to help understand how each term contributes to performance; (4) We obtain state-of-the-art performance on benchmark datasets.

\section{Related Work}
\par
Hashing methods can be categorized into data-independent and data-dependent methods, based on whether they are independent of training data. Representative data-independent methods include Locality Sensitive Hashing (LSH)~\cite{andoni2006near} and Shift-Invariant Kernels Hashing (SIKH)~\cite{raginsky2009locality}. Data-independent methods generally need longer hash codes for satisfactory performance, compared to data-dependent methods. Data-dependent methods can be further divided into unsupervised and supervised methods, based on whether supervised information is provided or not.

\par

Unsupervised hashing methods only utilize the training data points to learn hash functions, without using any supervised information. Notable examples for unsupervised hashing methods include Spectral Hashing (SH)~\cite{weiss2009spectral}, Binary Reconstructive Embedding (BRE)~\cite{kulis2009learning}, Iterative Quantization (ITQ)~\cite{gong2011iterative}, Isotropic Hashing(IsoHash)~\cite{kong2012isotropic}, graph-based hashing methods~\cite{liu2011hashing,liu2014discrete,jiang2015scalable} and two deep hashing methods: Semantic Hashing~\cite{salakhutdinov2009semantic} and the hashing method proposed in~\cite{erin2015deep}.

\par
Supervised hashing methods leverage labeled data to learn hash codes. Typically, the supervised information are provided in one of three forms: point-wise labels, pairwise labels or ranking labels~\cite{li2015feature}.
Representative point-wise label based methods include CCA-ITQ~\cite{gong2011iterative} and the deep hashing method proposed in~\cite{lin2015deep}. Representative pairwise label based hashing methods include Minimal Loss Hashing (MLH)~\cite{norouzi2011minimal}, Supervised Hashing with Kernels (KSH)~\cite{liu2012supervised}, Latent Factor Hashing (LFH)~\cite{zhang2014supervised}, Fash Supervised Hashing (FASTH)~\cite{lin2014fast} and two deep hashing methods: CNNH~\cite{xia2014supervised} and DPSH~\cite{li2015feature}. Representative ranking label based hashing methods include Ranking-based Supervised Hashing(RSH)~\cite{wang2013learning}, Column Generation Hashing (CGHASH)~\cite{li2013learning} and the deep hashing methods proposed in~\cite{zhao2015deep,lai2015simultaneous,zhang2015bit,erin2015deep}. One special case of ranking labels is triplet labels.


Most existing supervised hashing methods represent images as off-the-shelf visual features and perform hash code learning independent of visual features, including some deep hashing methods~\cite{salakhutdinov2009semantic,erin2015deep}. However, off-the-shelf features may not be optimally compatible with hash code learning procedure and thus results in sub-optimal hash codes.
\par
CNNH~\cite{xia2014supervised}, supervised by triplet labels, is the first proposed deep hashing method without using off-the-shelf features. However, CNNH cannot learn image features and hash codes simultaneously and still has limitations. This has been verified by the authors of CNNH themselves in a follow-up work~\cite{lai2015simultaneous}. Other ranking label or triplet label based deep hashing methods include Network in Network Hashing (NINH)~\cite{lai2015simultaneous}, Deep Semantic Ranking based Hashing (DSRH)~\cite{zhao2015deep}, Deep Regularized Similarity Comparison Hashing (DRSCH)~\cite{zhang2015bit} and Deep Similarity Comparison Hashing (DSCH)~\cite{zhang2015bit}. {While these methods can simultaneously perform image feature learning and hash code learning given supervision of triplet labels, we present a novel formulation of the likelihood of the given triplet labels to evaluate the quality of learned hash codes.
\par
Deep Pairwise-Supervised Hashing (DPSH)~\cite{li2015feature} is the first proposed deep hashing method to simultaneously perform image feature learning and hash code learning with pairwise labels and achieves highest performance compared to other deep hashing methods. Our method is supervised by triplet labels because triplet labels inherently contain richer information than pairwise labels.

\begin{figure}[t]
\begin{center}
\centering
\includegraphics[width=0.9\columnwidth]{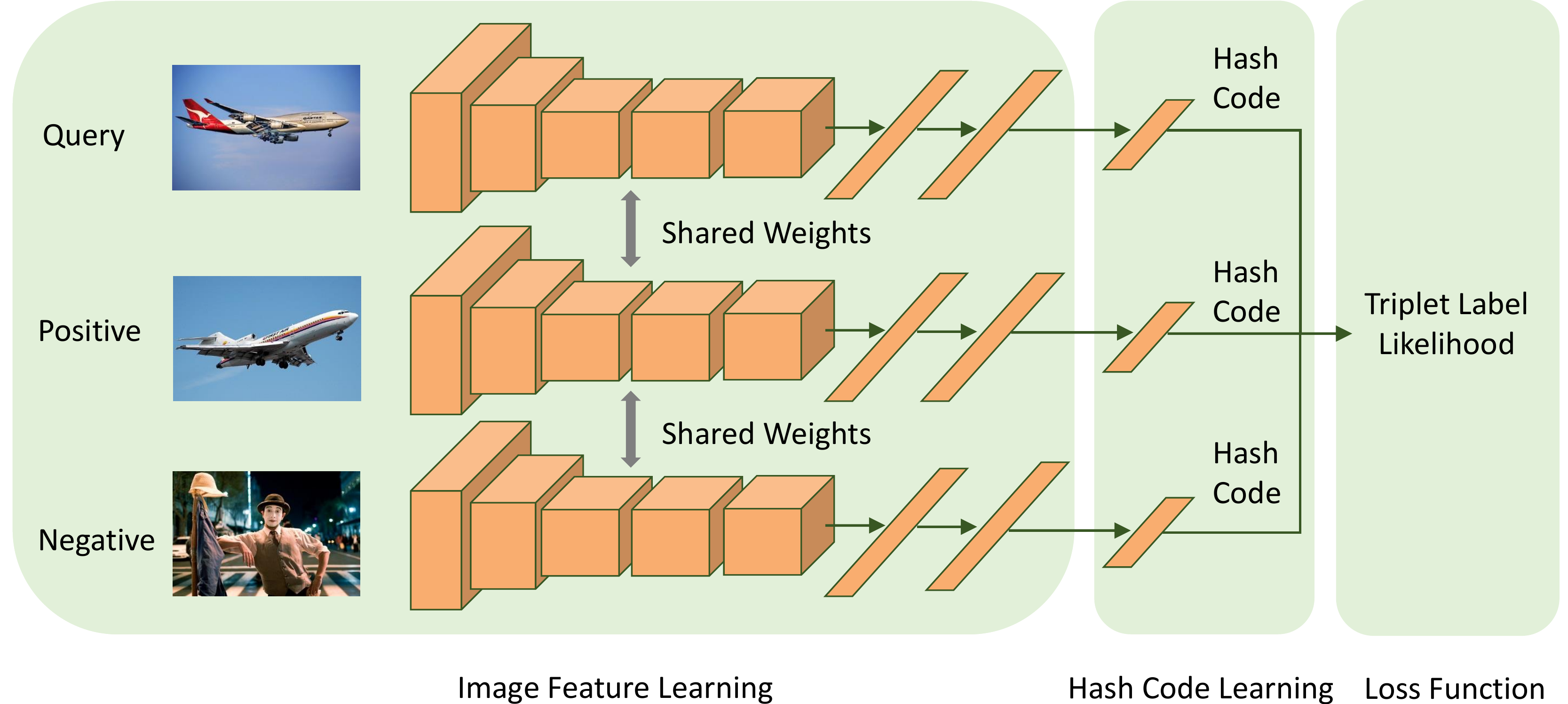}
\end{center}

\caption{Overview of the proposed end-to-end deep hashing method.}
\label{framework}
\end{figure}

\section{Approach}

We first give the formal definition of our problem and then introduce our proposed end-to-end method which simultaneously performs image feature learning and hash code learning from triplet labels.

\subsection{Problem Definition}

Given $N$ training images $\mathcal{I} = \{I_1, \ldots, I_N\}$ and $M$ triplet labels $\mathcal{T} = \{(q_1, p_1, n_1),\\ \ldots, (q_M, p_M, n_M)\}$ where the triplet of image indices $(q_m, p_m, n_m)$ denotes that the query image of index $q_m \in \{1 \ldots N\}$ is more similar to the positive image of index $p_m \in \{1 \ldots N\}$  than to the negative image of index $n_m \in \{1 \ldots N\}$. One possible way of generating triplet labels is by selecting two images from the same semantic category ($I_{q_m}$ and $I_{p_m}$) and selecting the negative ($I_{n_m}$) from a different semantic category.

Our goal is to learn a hash code $\mathbf{b}_n$ for each image $I_n$, where $\mathbf{b} \in \{+1,-1\}^{L}$ and $L$ is the length of hash codes. The hash codes $\mathcal{B} = \{\mathbf{b}_n \}_{n=1}^{N}$ should satisfy all the triplet labels $\mathcal{T}$ as much as possible in the Hamming space. More specifically, $\text{dist}_H(\mathbf{b}_{q_m}, \mathbf{b}_{p_m})$ should be smaller than $\text{dist}_H(\mathbf{b}_{q_m}, \mathbf{b}_{n_m})$ as much as possible, where $\text{dist}_H(\cdot, \cdot)$ denotes the Hamming distance between two binary codes.
Generally speaking, we aim to learn a hash function $h(\cdot)$ to map images to hash codes. {We can write $h(\cdot)$ as $[h_1(\cdot), ..., h_L(\cdot)]$} and for image $I_n$, its hash code can be denoted as $\mathbf{b}_n = h(I_n) = [h_1(I_n), ..., h_L(I_n)]$.

\subsection{Learning the Hash Function} 


Most previous hashing methods rely on off-the-shelf visual features, which may not be optimally compatible with the hash code learning procedure. Thus deep hashing methods~\cite{zhao2015deep,zhang2015bit,lai2015simultaneous,li2015feature} are proposed to simultaneously learn image features and hash codes from images. We propose a novel deep hashing method utilizing triplet labels, which simultaneously performs image feature learning and hash code learning in an end-to-end manner. As shown in Fig.~\ref{framework}, our method consists of three key components: (1) an image feature learning component, (2) a hash code learning component and (3) a loss function component.

\subsubsection{Image Feature Learning}

This component is designed to employ a Convolutional neural network to learn visual features from images. We adopt the CNN-F network architecture~\cite{chatfield2014return} for this component. CNN-F has eight layers, where the last layer is designed to learn the probability distribution over category labels. So only the first 7 layers of CNN-F are used in this component. Other networks like AlexNet~\cite{krizhevsky2012imagenet}, residual network~\cite{he2015deep} can also be used in this component.

\subsubsection{Hash Code Learning}

This component is designed to learn hash codes of images. We use one fully connected layer for this component and we want this layer to output hash codes of images. In particular, the number of neurons of this layer equals the length of targeted hash codes. Multiple fully connected layers or other architectures like the divide-and-encode module proposed by~\cite{lai2015simultaneous} can also be applied here. We do not focus on this in this work and leave this for future study.

\subsubsection{Loss Function}

This component measures how well the given triple labels are satisfied by the learned hash codes by computing the likelihood of the given triplet labels. Inspired by the likelihood of the pairwise similarities proposed in LFH~\cite{zhang2014supervised}, we present our formulation of the likelihood for a given triplet label. We call this the \textit{triplet label likelihood} throughout the text.

Let $\Theta_{ij}$ denote half of the inner product between two hash codes $\mathbf{b}_i, \mathbf{b}_j \in \{+1, -1\} ^ L$:
\begin{equation}
\Theta_{ij} = \frac{1}{2} \mathbf{b}_i ^T \mathbf{b}_j
\end{equation}
Then the triplet label likelihood is formulated as:
\begin{equation}
p(\mathcal{T}\mid \mathcal{B}) = \prod_{m=1}^{M} p((q_m, p_m, n_m) \mid \mathcal{B})
\end{equation}
with
\begin{equation}\label{eqprob}
p((q_m, p_m, n_m) \mid \mathcal{B}) = \sigma(\Theta_{q_m p_m} - \Theta_{q_m n_m} - \alpha)
\end{equation}
where $\sigma(x)$ is the sigmoid function $\sigma(x) = \frac{1}{1+e^{-x}}$, $\alpha$ is the margin, a positive hyper-parameter and $\mathcal{B}$ is the set of all hash codes.

We now show how maximizing the triplet label likelihood matches the goal to preserve the relative similarity between the query image, positive image and negative image. We first prove the following relationship between the Hamming distance between two binary codes and their inner product:
\begin{equation}\label{eqtheta}
\text{dist}_H(\mathbf{b}_{i}, \mathbf{b}_{j}) = \frac{1}{2}(L-2\Theta_{ij})
\end{equation}
According to Eq.~\ref{eqtheta}, we can have
\begin{equation}
\text{dist}_H(\mathbf{b}_{q_m}, \mathbf{b}_{p_m}) -\text{dist}_H(\mathbf{b}_{q_m}, \mathbf{b}_{n_m}) = -(\Theta_{q_mp_m} - \Theta_{q_mn_m})
\end{equation}
According to Eq.~\ref{eqprob}, we know that the larger $p((q_m, p_m, n_m) \mid \mathcal{B})$ is, the larger $(\Theta_{q_m p_m} - \Theta_{q_m n_m} - \alpha)$ will be. Since $\alpha$ is a constant number here, the larger $(\Theta_{q_m p_m} - \Theta_{q_m n_m} - \alpha)$ is, the smaller $(\text{dist}_H(\mathbf{b}_{q_m}, \mathbf{b}_{p_m}) -\text{dist}_H(\mathbf{b}_{q_m}, \mathbf{b}_{n_m}))$ will be. Thus, by maximizing the triplet label likelihood $p(\mathcal{T}\mid \mathcal{B})$, we can enforce the Hamming distance between the query and the positive image to be smaller than that between the query and the negative image. The margin $\alpha$ here can regularize the distance gap between  $\text{dist}_H(\mathbf{b}_{q_m}, \mathbf{b}_{p_m})$ and $\text{dist}_H(\mathbf{b}_{q_m}, \mathbf{b}_{n_m})$. The margin $\alpha$ can also help speed up training our model as explained later in this Section and verified in our experiments in Sec~\ref{ablation}.

\par
Now we define our loss function as the negative log triplet label likelihood as follows,

\begin{equation} \label{eqloss11}
\begin{split}
L &= -\log p(\mathcal{T}\mid \mathcal{B}) \\
&= -\sum_{m=1}^{M} \log p((q_m, p_m, n_m) \mid \mathcal{B}) \\
\end{split}
\end{equation}
Plug Eq.~\ref{eqprob} into the above equation, we can drive that
\begin{equation} \label{eqloss1}
L= -\sum_{m=1}^{M} (\Theta_{q_mp_m} - \Theta_{q_mn_m} - \alpha - \log(1+e^{\Theta_{q_mp_m} - \Theta_{q_mn_m} - \alpha}))
\end{equation}


\par
Minimizing the loss defined in (\ref{eqloss1}) is an intractable discrete optimization problem~\cite{li2015feature}. One can choose to relax $\{\mathbf{b}_{n}\}$ from discrete to continuous, \emph{i.e.}, relax $\{\mathbf{b}_{n}\}$ to $\{\mathbf{u}_{n}\}$, where $\mathbf{u}_{n} \in \mathbb{R}^{L\times 1}$ and then minimize the loss. This is the strategy employed by LFH~\cite{zhang2014supervised}, but this may be harmful to the performance due to the relaxation error~\cite{kang2016column}.  In the context of hashing, the relaxation error is actually the quantization error. Although optimal real vectors $\{\mathbf{u}_{n}\}$ are learned, we still need to quantize them to binary codes $\{\mathbf{b}_{n}\}$. This process induces quantization error. To reduce it, we propose to also consider the quantization error when solving the ralexed problem.
\par
Concretely, we relax binary codes $\{\mathbf{b}_{n}\}$ to real vectors $\{\mathbf{u}_{n}\}$ and re-define $\Theta_{ij}$ as
\begin{equation}
\Theta_{ij} = \frac{1}{2} \mathbf{u}_i ^T \mathbf{u}_j
\end{equation}
and our loss function becomes

\begin{equation} \label{eqloss2}
\begin{split}
L =&-\sum_{m=1}^{M} (\Theta_{q_mp_m} - \Theta_{q_mn_m} - \alpha - \log(1+e^{\Theta_{q_mp_m} - \Theta_{q_mn_m} - \alpha})) \\
&+ \lambda\sum_{n=1}^{N}||\mathbf{b}_{n} - \mathbf{u}_{n}||_2^{2}
\end{split}
\end{equation}
where $\lambda$ is a hyper-parameter to balance the negative log triplet likelihood and the quantization error and $\mathbf{b}_{n}=sgn(\mathbf{u}_{n})$, where $sgn(\cdot)$ is the sign function and $sgn(\mathbf{u}_{n}^{(k)})$ equals to 1 if $\mathbf{u}_{n}^{(k)} > 0$ and -1, otherwise. The quantization error term $\sum_{n=1}^{N}||\mathbf{b}_{n} - \mathbf{u}_{n}||_2^2$ is also adopted as a regularization term in DPSH~\cite{li2015feature}. However, they do not interpret it as quantization error.


\subsubsection{Model Learning}
The three key components of our method can be integrated into a Siamese-triplet network as shown in Fig.~\ref{framework}, which takes triplets of images as input and output hash codes of images. The network consists of three sub-networks with exactly the same architecture and shared weights. In our experiments, the sub-network is a fully connected layer on top of the first seven layers of CNN-F~\cite{chatfield2014return}. We train our network by minimizing the loss function:
\begin{equation}
\begin{split}
L(\theta) = &-\sum_{m=1}^{M} (\Theta_{q_mp_m} - \Theta_{q_mn_m} - \alpha - \log(1+e^{\Theta_{q_mp_m} - \Theta_{q_mn_m} - \alpha})) \\
&+ \lambda\sum_{n=1}^{N}||\mathbf{b}_{n} - \mathbf{u}_{n}||_2^{2}
\end{split}
\end{equation}
where $\theta$ denotes all the parameters of the sub-network, $\mathbf{u}_{n}$ is the output of the sub-network with $n^{th}$ training image and $\mathbf{b}_{n} = sgn(\mathbf{u}_{n})$. We can see that $L$ is differentiable with respect to $\mathbf{u}_{n}$. Thus the back-propagation algorithm can be applied here to minimize the loss function.
\par
Once training is completed, we can apply our model to generate hash codes for new images. For a new image $I$, we pass it into the trained sub-network and take the output of the last layer $\mathbf{u}$. Then the hash code $\mathbf{b}$ of image $I$ is $\mathbf{b} = sgn(\mathbf{u})$.

\subsubsection{Impact of Margin $\alpha$} We argue that a positive margin $\alpha$ can help speed up the training process. We now analyze this theoretically by looking at the derivative of the loss function. In particular, for $n^{th}$ image, we compute the derivative of the loss function $L$ with respect to $\mathbf{u}_n$ as follows:
\begin{equation} \label{eqderivative}
\begin{split}
\frac{\partial L}{\partial \mathbf{u}_n} = &-\frac{1}{2}\sum_{m:(n, p_m, n_m)\in \mathcal{T}} (1-\sigma(\Theta_{q_mp_m} - \Theta_{q_mn_m} - \alpha))(\mathbf{u}_{p_m} - \mathbf{u}_{n_m}) \\
&-\frac{1}{2}\sum_{m:(q_m, n, n_m)\in \mathcal{T}} (1-\sigma(\Theta_{q_mp_m} - \Theta_{q_mn_m} - \alpha)) \mathbf{u}_{q_m} \\
&+\frac{1}{2}\sum_{m:(q_m, p_m, n)\in \mathcal{T}} (1-\sigma(\Theta_{q_mp_m} - \Theta_{q_mn_m} - \alpha)) \mathbf{u}_{q_m} \\
&+ 2\lambda(\mathbf{u}_n - \mathbf{b}_n)
\end{split}
\end{equation}
where $\sigma(x)$ is the sigmoid function $\sigma(x) = \frac{1}{1+e^{-x}}$ and $\mathcal{T}$ is the set of triplet labels. In the derivative shown above, we observe this term $(1-\sigma(\Theta_{q_mp_m} - \Theta_{q_mn_m} - \alpha))$. We know that $\sigma(x)$ saturates very quickly, \emph{i.e.}, being very close to 1, as $x$ increases. If $\alpha = 0$, when $(\Theta_{q_mp_m} - \Theta_{q_mn_m})$ becomes positive, the term $(1-\sigma(\Theta_{q_mp_m} - \Theta_{q_mn_m} - \alpha))$ will be very close to 0. This will make the magnitude of the derivative very small and further make the model hard to train. A positive margin $\alpha$ adds a negative offset on $(\Theta_{q_mp_m} - \Theta_{q_mn_m})$ and can prevent $(1-\sigma(\Theta_{q_mp_m} - \Theta_{q_mn_m} - \alpha))$ from being very small. Further this makes the model easier to train and helps speed up the training process. We give experiment results to verify this in Sec~\ref{ablation}.

\section{Experiment}

\subsection{Datasets and Evaluation Protocol}
We conduct experiments on two widely used benchmark datasets, CIFAR-10~\cite{krizhevsky2009learning} and NUS-WIDE~\cite{chua2009nus}. The CIFAR-10~\cite{krizhevsky2009learning} dataset contains 60,000 color images of size $32 \times 32$, which can be divided into 10 categories and 6,000 images for each category. Each image is only associated with one category. The NUS-WIDE~\cite{chua2009nus} dataset contains nearly 27,000 color images from the web. Different from CIFAR-10, NUS-WIDE is a multi-label dataset. Each image is annotated with one or multiple labels in 81 semantic concepts. Following the setting in~\cite{xia2014supervised,lai2015simultaneous,zhang2015bit,li2015feature}, we only consider images annotated with 21 most frequent labels. For each of the 21 labels, at least 5,000 images are annotated with the label. In addition, NUS-WIDE provides links to images for downloading and some links are now invalid. This causes some differences between the image set used by previous work and our work. In total, we use 161,463 images from the NUS-WIDE dataset.
\par
We employ mean average precision (MAP) to evaluate the performance of our method and baselines similar to most previous work~\cite{xia2014supervised,lai2015simultaneous,zhang2015bit,li2015feature}. Two images in CIFAR-10 are considered similar if they belong to the same category. Two images in NUS-WIDE are considered similar if they share at least one label.

\subsection{Baselines and Setting}

Following~\cite{li2015feature}, we consider the following baselines:
\begin{enumerate}
\item Traditional unsupervised hashing methods using hand-crafted features, including SH~\cite{weiss2009spectral}, ITQ~\cite{gong2011iterative}.
\item Traditional supervised hashing methods using hand-crafted features, including SPLH~\cite{wang2010sequential}, KSH~\cite{liu2012supervised}, FastH~\cite{lin2014fast}, LFH~\cite{zhang2014supervised} and SDH~\cite{shen2015supervised}.
\item The above traditional hashing methods using features extracted by CNN-F network~\cite{chatfield2014return} pre-trained on ImageNet.
\item Pairwise label based deep hashing methods: CNNH~\cite{xia2014supervised} and DPSH~\cite{li2015feature}.
\item Triplet label based deep hashing methods: NINH~\cite{lai2015simultaneous}, DSRH~\cite{zhao2015deep}, DSCH~\cite{zhang2015bit} and DRSCH~\cite{zhang2015bit}.
\end{enumerate}

\par
When using hand-crafted features, we use a 512-dimensional GIST descriptor~\cite{oliva2001modeling} to represent CIFAR-10 images. For NUS-WIDE images, we reprenst them by a 1134-dimensional feature vector, which is the concatenation of a 64-D color histogram, a 144-D color correlogram, a 73-D edge direction histogram,a 128-D wavelet texture, a 225-D block-wise color moments and a 500-D BoW representation based on SIFT descriptors.

\par
Following~\cite{zhao2015deep}~\cite{li2015feature}, we initialize the first seven layers of our network with the CNN-F network~\cite{chatfield2014return} pre-trained on ImageNet. In addition, the hyper-parameter $\alpha$ is set to half of the length of hash codes, \emph{e.g.}, 16 for 32-bit hash codes and the hyper-parameter $\lambda$ is set to 100 unless otherwise stated.

\par
We compare our method to most baselines under the following experimental setting. Following~\cite{xia2014supervised,lai2015simultaneous,li2015feature}, in CIFAR-10, 100 images per category, \emph{i.e.}, in total 1,000 images, are randomly sampled as query images. The remaining images are used as database images. For unsupervised hashing methods, all the database images are used as training images. For supervised hashing methods, 500 database images per category, \emph{i.e.}, in total 5,000 images, are randomly sampled as training images. In NUS-WIDE, 100 images per label, \emph{i.e.}, in total 2,100 images, are randomly sampled as query images. Likewise, the remaining images are used as database images. For unsupervised hashing methods, all the database images are used as training images. For supervised hashing methods, 500 database images per label, \emph{i.e.}, in total 10,500 images, are randomly sampled as training images. Since NUS-WIDE contains a huge number of images, when computing MAP for NUS-WIDE, only the top 5,000 returned neighbors are considered.

\par
We also compare our method to DSRH~\cite{zhao2015deep}, DSCH~\cite{zhang2015bit}, DRSCH~\cite{zhang2015bit} and DPSH~\cite{li2015feature} under a different experimental setting. In CIFAR-10, 1,000 images per category, \emph{i.e}., in total 10,000 images, are randomly sampled as query images. The remaining images are used as database images and all the database images are used as training images. In NUS-WIDE, 100 images per label, \emph{i.e.}, in total 2,100 images, are randomly sampled as query images. The remaining images are used as database images and still, all the database images are used as training images. Under this setting, when computing MAP for NUS-WIDE, we only consider the top 50,000 returned neighbors.

\subsection{Performance Evaluation}

\begin{table}[t]
\centering
\caption{Mean Average Precision (MAP) under the first experimental setting. The MAP for NUS-WIDE is computed based on the top 5,000 returned neighbors. The best performance is shown in boldface. DPSH* denotes the performance we obtain by running the code provided by the authors of DPSH in our experiments.}
\label{table1}
\begin{tabular}{|l|cccc|l|cccc|}
\hline
\multirow{2}{*}{Method} & \multicolumn{4}{c|}{CIFAR-10} & \multicolumn{1}{l|}{\multirow{2}{*}{Method}} & \multicolumn{4}{c|}{NUS-WIDE} \\ \cline{2-5} \cline{7-10}
 & 12 bits & 24 bits & 32 bits & 48 bits & \multicolumn{1}{l|}{} & 12 bits & 24 bits & 32 bits & 48 bits \\ \hline
Ours & 0.710 & \textbf{0.750} & \textbf{0.765} & \textbf{0.774} & Ours & \textbf{0.773} & \textbf{0.808} & \textbf{0.812} & \textbf{0.824} \\ \hline
DPSH & \textbf{0.713} & 0.727 & 0.744 & 0.757 & DPSH* & 0.752 & 0.790 & 0.794 & 0.812 \\ \hline
NINH & 0.552 & 0.566 & 0.558 & 0.581 & NINH & 0.674 & 0.697 & 0.713 & 0.715 \\ \hline
CNNH & 0.439 & 0.511 & 0.509 & 0.522 & CNNH & 0.611 & 0.618 & 0.625 & 0.608 \\ \hline
FastH & 0.305 & 0.349 & 0.369 & 0.384 & FastH & 0.621 & 0.650 & 0.665 & 0.687 \\ \hline
SDH & 0.285 & 0.329 & 0.341 & 0.356 & SDH & 0.568 & 0.600 & 0.608 & 0.637 \\ \hline
KSH & 0.303 & 0.337 & 0.346 & 0.356 & KSH & 0.556 & 0.572 & 0.581 & 0.588 \\ \hline
LFH & 0.176 & 0.231 & 0.211 & 0.253 & LFH & 0.571 & 0.568 & 0.568 & 0.585 \\ \hline
SPLH & 0.171 & 0.173 & 0.178 & 0.184 & SPLH & 0.568 & 0.589 & 0.597 & 0.601 \\ \hline
ITQ & 0.162 & 0.169 & 0.172 & 0.175 & ITQ & 0.452 & 0.468 & 0.472 & 0.477 \\ \hline
SH & 0.127 & 0.128 & 0.126 & 0.129 & SH & 0.454 & 0.406 & 0.405 & 0.400 \\ \hline
\end{tabular}
\end{table}

\begin{table}[t]
\centering
\caption{Mean Average Precision (MAP) under the first experimental setting. The MAP for NUS-WIDE is computed based on the top 5,000 returned neighbors.  The best performance is shown in boldface.}
\label{table2}
\begin{tabular}{|l|cccc|cccc|}
\hline
\multirow{2}{*}{Method} & \multicolumn{4}{c|}{CIFAR-10} & \multicolumn{4}{c|}{NUS-WIDE} \\ \cline{2-9}
 & 12 bits & 24 bits & 32 bits & 48 bits & 12 bits & 24  bits& 32 bits & 48 bits \\ \hline
Ours & \textbf{0.710} & \textbf{0.750} & \textbf{0.765} & \textbf{0.774} & 0.773 & \textbf{0.808} & 0.812 & 0.824 \\ \hline
FastH + CNN & 0.553 & 0.607 & 0.619 & 0.636 & {0.779} & 0.807 & \textbf{0.816} & \textbf{0.825} \\ \hline
SDH + CNN & 0.478 & 0.557 & 0.584 & 0.592 & \textbf{0.780} & 0.804 & 0.815 & 0.824 \\ \hline
KSH + CNN & 0.488 & 0.539 & 0.548 & 0.563 & 0.768 & 0.786 & 0.79 & 0.799 \\ \hline
LFH + CNN & 0.208 & 0.242 & 0.266 & 0.339 & 0.695 & 0.734 & 0.739 & 0.759 \\ \hline
SPLH + CNN & 0.299 & 0.33 & 0.335 & 0.33 & 0.753 & 0.775 & 0.783 & 0.786 \\ \hline
ITQ + CNN & 0.237 & 0.246 & 0.255 & 0.261 & 0.719 & 0.739 & 0.747 & 0.756 \\ \hline
SH + CNN & 0.183 & 0.164 & 0.161 & 0.161 & 0.621 & \multicolumn{1}{c|}{0.616} & 0.615 & 0.612 \\ \hline
\end{tabular}
\end{table}

\begin{table}[t]
\centering
\caption{Mean Average Precision (MAP) under the second experimental setting. The MAP for NUS-WIDE is computed based on the top 50,000 returned neighbors. The best performance is shown in boldface. DPSH* denotes the performance we obtain by running the code provided by the authors of DPSH in our experiments.}
\label{table3}
\begin{tabular}{|l|cccc|cccc|}
\hline
\multirow{2}{*}{Method} & \multicolumn{4}{c|}{CIFAR-10} & \multicolumn{4}{c|}{NUS-WIDE} \\ \cline{2-9}
 & 16 bits & 24 bits & 32 bits & 48 bits & 16 bits & 24 bits & 32 bits & 48 bits \\ \hline
Ours & \textbf{0.915} & \textbf{0.923} & \textbf{0.925} & \textbf{0.926} & \textbf{0.756} & \textbf{0.776} & \textbf{0.785} & \textbf{0.799} \\ \hline
DPSH & 0.763 & 0.781 & 0.795 & 0.807 & 0.715 & 0.722 & 0.736 & 0.741 \\ \hline
DRSCH & 0.615 & 0.622 & 0.629 & 0.631 & 0.618 & 0.622 & 0.623 & 0.628 \\ \hline
DSCH & 0.609 & 0.613 & 0.617 & 0.62 & 0.592 & 0.597 & 0.611 & 0.609 \\ \hline
DSRH & 0.608 & 0.611 & 0.617 & 0.618 & 0.609 & 0.618 & 0.621 & 0.631 \\ \hline
DPSH* & 0.903 & 0.885 & 0.915 & 0.911 & \multicolumn{4}{c|}{N/A} \\ \hline
\end{tabular}
\end{table}

\subsubsection{Comparison to Traditional Hashing Methods using Hand-crafted Features} As shown in Table~\ref{table1}, we can see that on both datasets, our method outperforms previous hashing methods using hand-crafted features significantly. In Table~\ref{table1}, the results of NINH, CNNH, KSH and ITQ are from~\cite{xia2014supervised,lai2015simultaneous} and the results of other methods except our method are from~\cite{li2015feature}. This is reasonable as we use the same experimental setting and evaluation protocol.

\subsubsection{Comparison to Traditional Hashing Methods using Deep Features}
When we train our model, we initialize the first 7 layers of our network with CNN-F network~\cite{chatfield2014return} pre-trained on ImageNet. Thus one may argue that the boost in performance comes from that network instead of our method. To further validate our method, we compare our method to traditional hashing methods using deep features extracted by CNN-F network. As shown in Table~\ref{table2}, we can see that our method can significantly outperform traditional methods on CIFAR-10 and obtain comparable performance with the best performing traditional methods on NUS-WIDE. The results in Table~\ref{table2} are copied from~\cite{li2015feature}, which is reasonable as we used the same experimental setting and evaluation protocol.

\subsubsection{Comparison to Deep Hashing Methods} Now we compare our method to other deep hashing methods. In particular, we compare our method to CNNH, NINH and DPSH under the first experimental setting in Table~\ref{table1} and DSRH, DSCH, DRSCH and DPSH under the second experimental setting in Table~\ref{table3}. The results of DSRH, DSCH and DRSCH are directly from~\cite{zhang2015bit}. We can see that our method significantly outperforms all previous triplet label based deep hashing methods, including NINH, DSRH, DSCH and DRSCH.

We now compare our method to the current state-of-the-art method DPSH under the first experimental setting. As shown in Table~\ref{table1}, our method outperforms DPSH by about $2\%$ on both CIFAR-10 and NUS-WIDE datasets. Note that on NUS-WIDE, we are comparing to DPSH* instead of DPSH. DPSH represents the performance reported in~\cite{li2015feature} and DPSH* represents the performance we obtain by running the code of DPSH provided by the authors of~\cite{li2015feature} on NUS-WIDE. We re-run their code on NUS-WIDE because the NUS-WIDE dataset does not provide the original images to download instead of the links to image, which results in some differences between the images used by them~\cite{li2015feature} and us.

As shown in Table~\ref{table3}, our method outperforms DPSH by more than $10\%$ on CIFAR-10 and about $5\%$ on NUS-WIDE under the second experimental setting. We also re-run the code of DPSH on CIFAR-10 under the same setting and we can obtain much higher the performance then what was reported in~\cite{li2015feature}.\footnote{We communicated with the authors of DPSH. The main difference between our experiments and their experiments is the step size and decay factor for learning rate change. They say that with our parameters, they can also get better results than what is reported in their paper~\cite{li2015feature}.} The performance we obtain is denoted by DPSH* in Table~\ref{table3} and we can see our method still outperforms DPSH* by about $1\%$. We also show some retrieval examples on NUS-WIDE with hash codes learned by our method and DPSH respectively in Fig.~\ref{visexample}.

\subsection{Ablation Studies}
\subsubsection{Impact of the hyper-parameter $\alpha$}
\label{ablation}
Figure~\ref{alpha} shows the effect of the margin $\alpha$ at 32 bits on CIFAR-10 dataset. We can see that within the same number of training epochs, we can obtain better performance with a larger margin. This verifies our previous analysis in Sec 3.

\subsubsection{Impact of the hyper-parameter $\lambda$}

Figure~\ref{lambda} shows the effect of the hyper-parameter $\lambda$ in Eq.~\ref{eqloss2} at 12 bits and 32 bits on CIFAR-10 dataset. As one can see, for both 12 bits and 32 bits, there is a significant performance drop in terms of MAP when $\lambda$ becomes very small (\emph{e.g.}, 0.1) or very large (\emph{e.g.}, 1000). This is reasonable since $\lambda$ is designed to balance the negative log triplet likelihood and the quantization error. Setting $\lambda$ to small or to large will lead to inbalance between these two terms.

\subsubsection{Impact of the number of training images} We also study the impact of the number of training images on the performance. Figure~\ref{training} shows the performance of our method at 12 bits on CIFAR-10 using different number of training images. We can see that more training images will incur noticeable improvements.
\begin{figure}[p]
\begin{center}
\centering
\includegraphics[width=0.95\columnwidth]{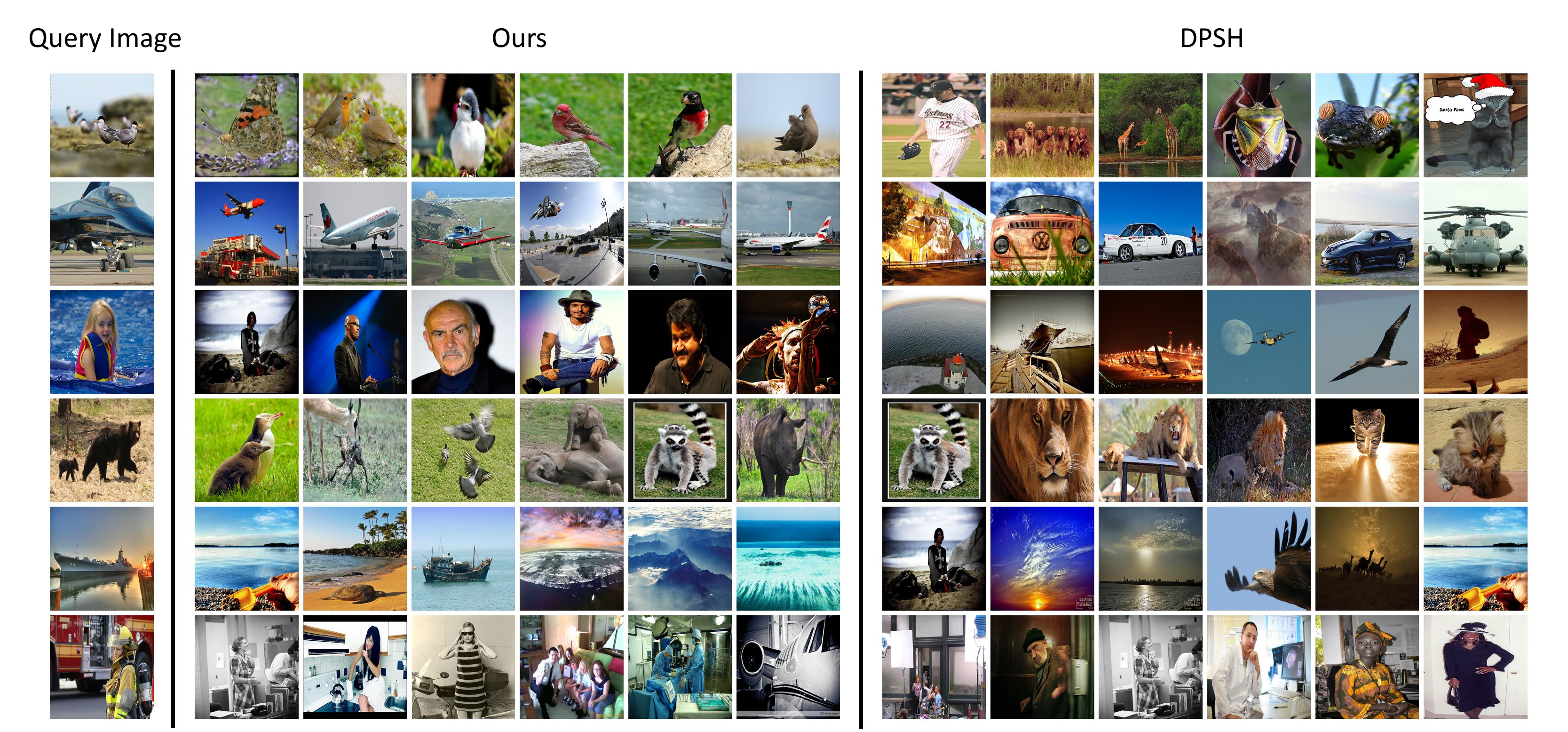}
\end{center}
\caption{Retrieval Examples. Left: Query images. Middle: Top images retrieved by hash codes learnt by our method. Right: Top images retrieved by hash codes learnt by DPSH~\cite{li2015feature}.}
\label{visexample}
\end{figure}

\begin{figure*}[p]
\centering
\subfloat[Impact of $\alpha$]{
\begin{minipage}{0.45\linewidth}
\centering
\includegraphics[width = 1.0\columnwidth]{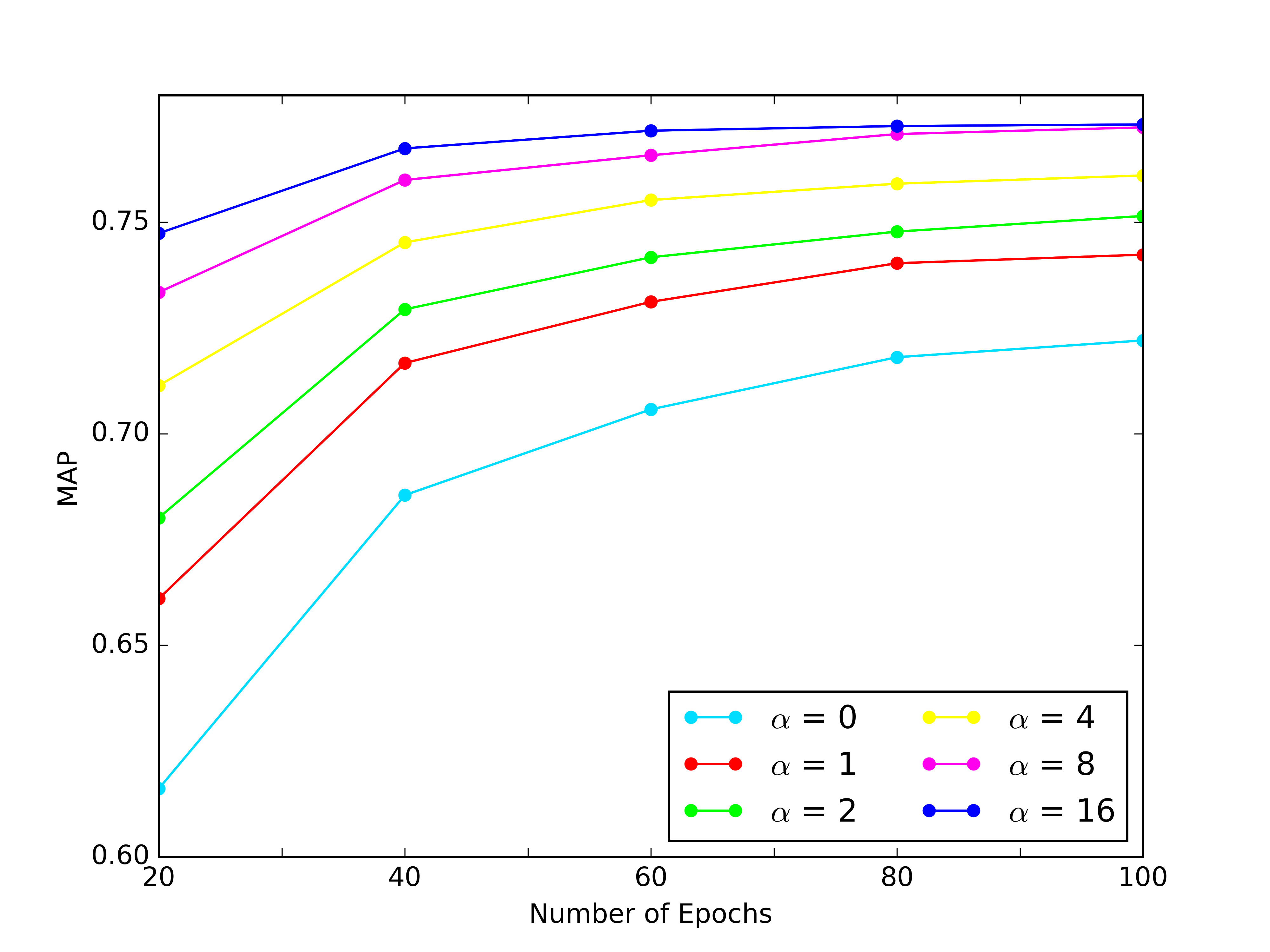}
\label{alpha}
\end{minipage}
}
\hfill
\subfloat[Impact of $\lambda$]{
\begin{minipage}{0.45\linewidth}
\centering
\includegraphics[width = 1.0\columnwidth]{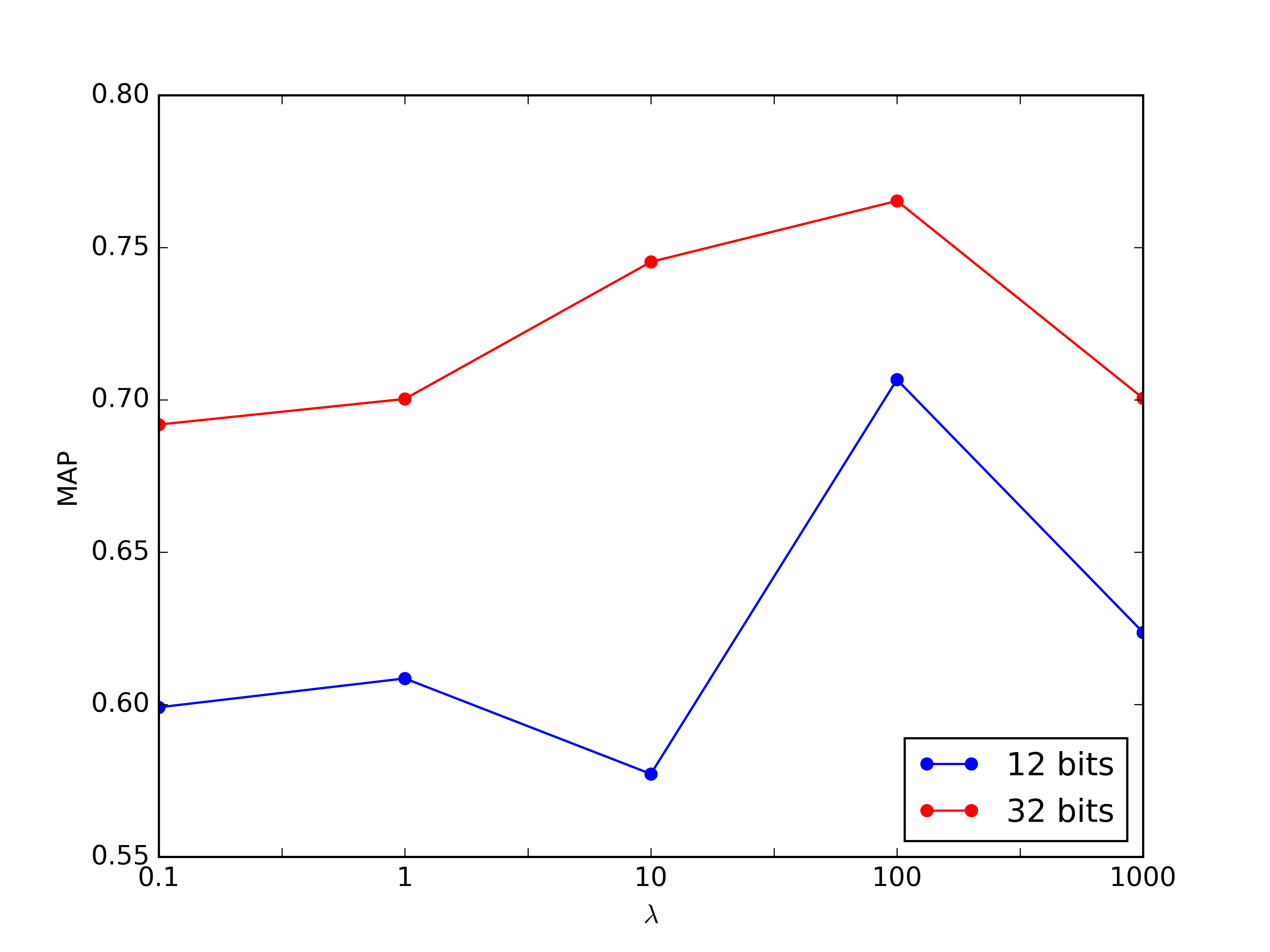}
\label{lambda}
\end{minipage}
}
\hfill
\subfloat[Impact of number of training images]{
\begin{minipage}{0.45\linewidth}
\centering
\includegraphics[width = 1.0\columnwidth]{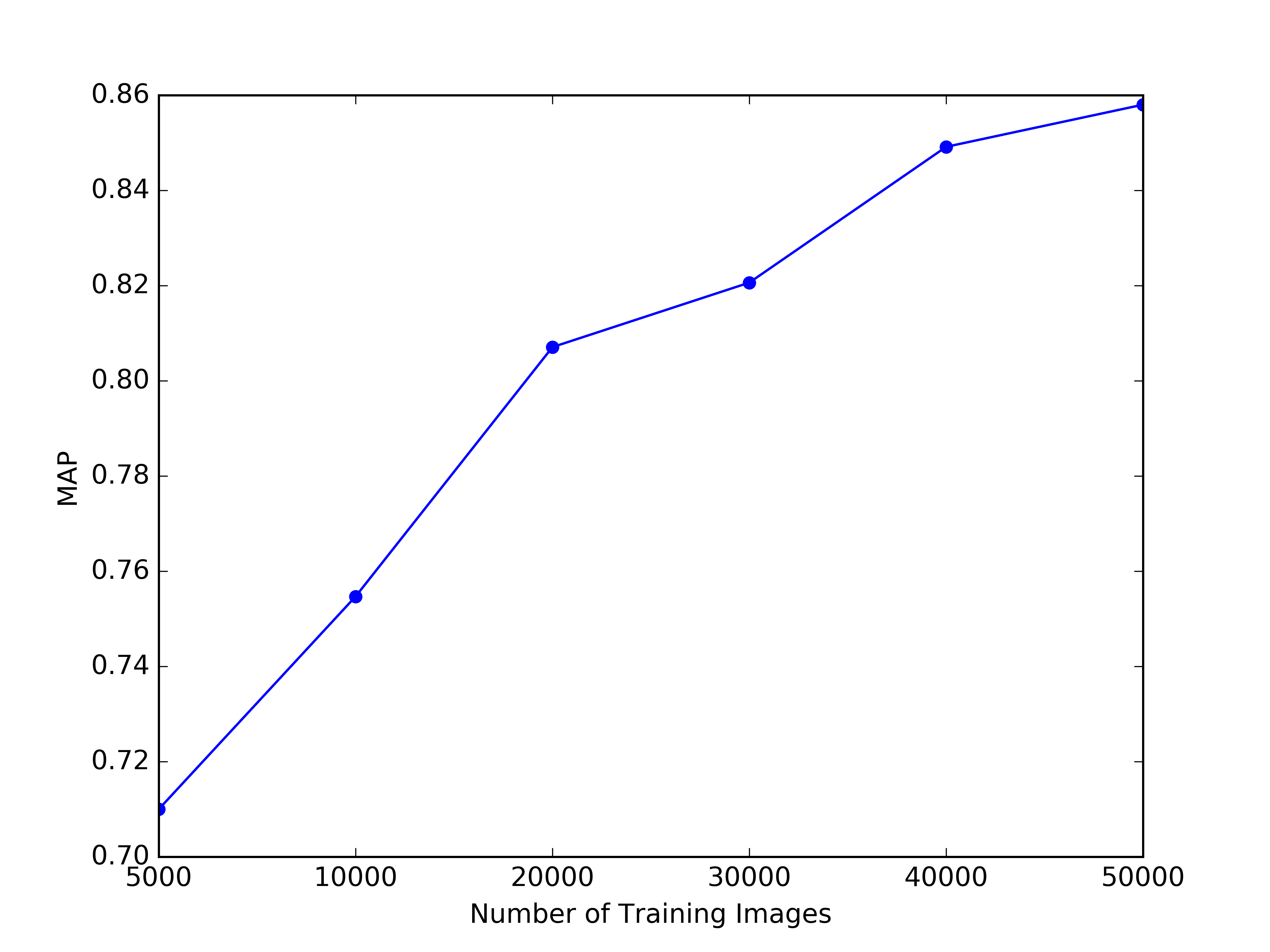}
\label{training}
\end{minipage}
}
\caption{Ablation Studies.}
\end{figure*}


\section{Conclusion}
In this paper, we have proposed a novel deep hashing method to simultaneously learn image features and hash codes given the supervision of triplet labels. Our method learns high quality hash codes by maximizing the likelihood of given triplet labels under learned hash codes. Extensive experiments on standard benchmark datasets show that our method outperforms all the baselines, including the state-of-the-art method DPSH~\cite{li2015feature} and all the previous triplet label based deep hashing methods.


\vspace{3mm}
\noindent {\bf Acknowledgement}. This work was sponsored by DARPA under agreement number FA8750-14-2-0244. The U.S. Government is authorized to reproduce and distribute reprints for Governmental purposes notwithstanding any copyright notation thereon. The views and conclusions contained herein are those of the authors and should not be interpreted as necessarily representing the official policies or endorsements, either expressed or implied, of DARPA or the U.S. Government.
\bibliographystyle{splncs}
\bibliography{egbib}


\end{document}